\documentclass{article}
\usepackage{spconf}
\usepackage[utf8]{inputenc} 
\usepackage[T1]{fontenc}    
\usepackage{hyperref}       
\usepackage{url}            
\usepackage{booktabs}       
\usepackage{amsfonts}       
\usepackage{nicefrac}       
\usepackage{graphicx}
\usepackage{tikz}
\usepackage{tikz-qtree}

\renewcommand{\paragraph}{%
  \@startsection{paragraph}{4}{\z@}%
                {1.5ex \@plus 0.5ex \@minus 0.2ex}%
                {-1em}%
                {\normalsize\bf}%
}

\usepackage{fancyhdr}
\title{Parsing Coordination for Spoken Language Understanding}
\name{Sanchit Agarwal, Rahul Goel, Tagyoung Chung, Abhishek Sethi, Arindam Mandal, Spyros Matsoukas
}

\address{\tt{\{agsanchi, goerahul, tagyoung, abhsethi, arindamm, matsouka\}@amazon.com}}

\begin{document}

\maketitle

\thispagestyle{fancy}

\begin{abstract}

Typical spoken language understanding systems provide narrow semantic
parses using a domain-specific ontology. The parses contain intents and slots
that are directly consumed by downstream domain applications. In this work we
discuss expanding such systems to handle compound entities and intents by
introducing a domain-agnostic shallow parser that handles linguistic coordination.
We show that our model for parsing coordination learns
domain-independent and slot-independent features and is able to segment conjunct
boundaries of many different phrasal categories. We also show that using
adversarial training can be effective for improving generalization across
different slot types for coordination parsing.

\end{abstract}


\section{Introduction}\label{sec:intro}

A typical spoken language understanding (SLU) system maps user utterances to
domain-specific semantic representations that can be factored into an intent and
slots~\cite{mesnil2015using,xu2013convolutional}. For example, an utterance,
{\it ``what is the weather like in boston'' } has one intent WeatherInfo and one
slot type CityName whose value is {\it ``boston.''} Thus, parsing for such
systems is often factored into two separate tasks: intent classification and
entity recognition whose results are consumed by downstream domain applications.
Such domain-specific parser is an example of narrow
parser~\cite{allen2018effective}.  It produces detailed domain-specific
information to meet the needs of an application but cannot handle utterances
outside of the domain or general linguistic structures well. Many of such
parsers also impose rigid constraints such as allowing only one intent to be
associated with an utterance and only one value to be associated per slot type.
In this paper we discuss a way to enable understanding of compound entities and
intents by augmenting an SLU system with a domain-agnostic parser that can
handle coordination structure. This approach has the benefit of not having to
disrupt operational SLU services because it does not need to fundamentally
change underlying system or associated domain-specific parsers.

Compound intents and slots are typically expressed in an utterance using
linguistic coordination. For example, an utterance, {\it ``add peanut butter and
jelly to my list''} contains one coordination where coordinating conjunction is
{\it ``and''} and conjuncts are {\it ``peanut butter''} and {\it``jelly.''}
Although detecting coordinating structure can be solved by off-the-shelf broad
syntactic parsers, many of them are trained on written text with correct
punctuations. Automatic Speech Recognition (ASR) outputs often lack
punctuations, moreover, spoken language is often divergent from written
language. Therefore, such general-purpose syntactic parsers are not able to
accurately parse coordinating structure over varied phrase categories of spoken
language, which we discuss in the later section in more detail. By training a
shallow parser on spoken data on varied coordination structures, we can
correctly parse conjunctions and conjuncts in spoken utterances. We specifically
approach the problem as that of chunking~\cite{abney1991parsing}, where our
parser produces constituent boundaries of coordination structure. Once such
information is available, we can augment the original SLU system in multiple
different ways. One approach would be reformulating original utterances into
multiple simple utterances that downstream domain applications can handle.
Another way is to augment slots that original domain-specific parsers found with
the conjunct boundary information from the shallow coordination parser.

Instead of building multiple coordination parsers for different slot types or
domains, we build one parser that is capable of parsing coordination of multiple
different slot types, which are often of different phrase categories.  For
example, the utterance {\it ``add peanut butter and jelly to my list''} has
coordination of noun phrases (NN/NP) and {\it ``add do homework and call mom to
my to do list''} has coordination of full clauses (S/SBAR). We show that
training a DNN-based chunking model that is trained on a corpus that contains
coordination of varied slot types of different phrase categories, we can build a
coordination parser that performs well across different slot types. We also show
that using slot type information as adversarial loss enables us to train a model
that generalizes better on unseen slot types.

Our contributions in this paper are as follows. First, we show that a popular
off-the-shelf syntactic parser can be ineffective at correctly parsing
coordination in spoken language utterances. Second, we show that a simple
DNN-based chucking model is sufficient to parse varied coordination of different
slot types. We try different model architectures and report memory, latency, and
accuracy trade-off in different scenarios.  Finally, we show how to incorporate
adversarial loss while training such parsers so as to improve their
generalization on unseen data.

\section{Models}\label{sec:models}
Shallow parsing (chunking) problem is generally solved as sequence tagging
task~\cite{tjong2000introduction, ramshaw1999text}. We experiment with various sequence tagging DNN architectures with different configurations, all of which are based on bi-directional LSTM models. ~\cite{graves2013hybrid}.
There are three major ways that the models we experimented with were different from each other:

\begin{enumerate}

\item A character-level encoder: This component extracts a feature vector for each word from its characters. For each word $i$, we use bi-directional LSTMs to extract
character-level features with the intuition that they will inform the model about morphology of words. To make our character-level features fixed length, we concatenate the last state of forward LSTM with the first state of backward LSTM~\cite{ling2015not}. This gives us a character-based representation of the
word $w_i^{char}$. We chose to have character-level encoder for some experiments and we left it out for other experiments.

\item A word-level encoder: This component extracts features from surrounding words. When used together with character-level encoder, we concatenate the character-level embedding of the word with its word embedding: $w_i^{full} = [w_i^{char}; w_i^{emb}]$.
When only word embeddings are used, word-level encoder only uses word embedding: $w_i^{full} = [w_i^{emb}]$. When only character-level embeddings are used, word-level encoder only uses the character-level embedding of a word: $w_i^{full} = [w_i^{char}]$. Given an utterance of length $l$ and $w_1^{full}, w_2^{full}...w_l^{full}$, we use bi-directional LSTMs to get hidden representations $h_1, h_2...h_l$. We chose to initialize word embeddings with pre-trained embeddings for some experiments and we randomly initialized them for other experiments. 

\item A tag decoder: This component outputs a probability distribution over tags
given a current encoded state. For sequence tagging problem, the decoder
typically consists of a linear dense layer that transforms the encoded state
into the output space.  More specifically, for each word $i=1...l$, it computes
the output as $y_i = h_i*W + b$, where $W$, $b$ are the weight matrix and biases of the dense layer respectively. To convert the raw scores into probabilities, softmax normalization is typically applied and the network parameters are learned by optimizing the cross entropy loss. We also experimented with replacing the softmax cross entropy layer in the decoder with a CRF layer as described in the next paragraph.

\end{enumerate}

\begin{figure}
  \begin{center}
    \scalebox{1.06}{
    \includegraphics[width=8cm]{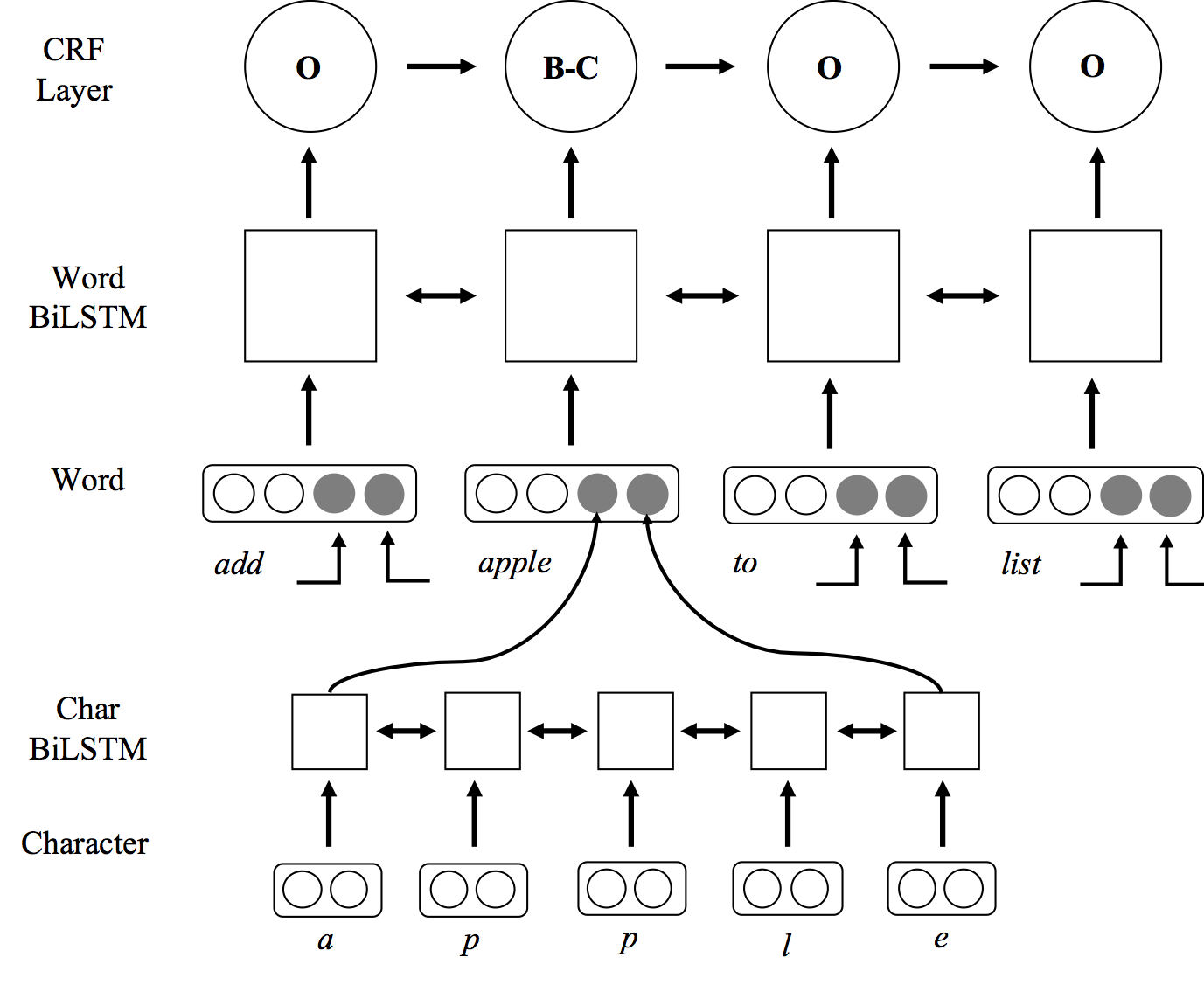}}
    \end{center}
  \caption{A model with character bi-directional LSTM, LSTM tagger, and a CRF
layer. It shows how we get word representation when character-level LSTM is
present. Word representation is obtained by concatenating word embedding with
character-level LSTM encoded character-level embedding. The word representation
is then fed the word-level LSTM. We also experiment with using only the
character-level encoder or only the word-level encoder for word representation.}
  \label{fig:model}
\end{figure}

Figure~\ref{fig:model} shows a network with word representation, a word-level LSTM encoder, and a CRF layer for tag decoder. It also explains how word representation is formed in the presence of a character-level LSTM. Note that in some of our experiment settings, only one of the character-level based embedding or word embedding is used to obtain word vector representation.

\vspace{1.0ex}
{\bf LSTM-CRF Networks:} In conditional random field
(CRF)~\cite{lafferty2001conditional}, it is possible to make use of the information from neighboring labels to predict the current label by incorporating the label transition information into the sequence labeling model. By augmenting LSTM with this property of CRF, the neural network can effectively use the past input features produced by a LSTM layer and combine them with sentence-level label information using a CRF layer~\cite{lample2016neural}. A CRF layer has state transition matrix as parameters where the states are the
output label space. For a sequence $x_{1,2..T}$ and corresponding tags $y_{1,2...T}$, let $f(x_{1,2...T})$ be the output of the bi-directional LSTM. The element $f_{i,t}$ is the score for tag $i$ at the time step $t$. The CRF layer introduces a transition matrix $A$ where each element $A_{i,j}$ is the score of transition from $i$-th state to $j$-th state for adjacent time steps. Note that this transition matrix is position independent. The score of the sentence $x_{1,2..T}$ and corresponding tags $y_{1,2...T}$ is given by a sum of transition and network scores:
\begin{equation}
  s(x_{1,2..T}, y_{1,2..T}) = \sum^T_{t=1}(A_{y_{t-1},y_t} + f_{y_t,t})
\end{equation}
To compute the parameters of $A$, dynamic programming is employed whose time complexity is $O(TE^2)$ where $E$ is the number of output tag types~\cite{collobert2011natural}. At inference time we use the Viterbi algorithm~\cite{forney1973viterbi} to compute the optimal label sequence.

\vspace{1.0ex}
{\bf Adversarial Training:} We want our model to perform well across different
slot types and possibly on unseen slot types, i.e., we want our models to learn
slot-type invariant features. In order to do this, we use techniques proposed
by~\cite{ganin2014unsupervised}, where the authors learned domain independent
features for domain adaptation. In our case, domain is analogous to slot types
that are coordinated.
In this approach, we augment previously described bi-directional LSTM networks with a feed-forward network that predicts domain with a gradient reversal layer during training, whose purpose is to make the network discriminate for the main learning task (sequence tagging) while keeping it invariant to the shift between domains. Given a learning target $y$ and a domain $d$ that we want the model to be invariant against, we minimize the loss $L_y$ (loss associated with sequence tagging) while maximizing the loss $L_d$ (loss associated with domain discrimination). This is achieved using a setup similar to multitasking DNNs such as~\cite{klerke2016improving} where the first few layers extract common features, which are fed to task-specific and domain-specific layers. Let $\theta_f$ be the parameters of the common layers, $\theta_y$ be the parameters of task-specific layer and $\theta_d$ be the parameters of domain-specific layers. For each data point $i=1..N$, the error function we are trying to minimize is
\begin{equation}
  E(\theta_f,\theta_y,\theta_d) = \sum_{i=1..N}(L_y^i(\theta_f, \theta_y) \ -
  \lambda \cdot L_d^i(\theta_f, \theta_d))
\end{equation}
where $\lambda$ is a hyper-parameter controlling the weightage on the adversarial component in the model.

During training $\lambda$ is varied as suggested in ~\cite{ganin2016domain}, 
\begin{equation}
\lambda_p = \frac{2}{1 + exp(-\gamma \cdot p) - 1}
\end{equation} 
where $p$ is the training progress and $\gamma$ controls how quickly we increase the weight on the adversarial component in the loss. We want to slowly increase $\lambda$ as we want the slot classifier to be less sensitive to noisy signal at the early stages of the training.

\section{Experiments}\label{sec:exp}

\subsection{Data}\label{sec:data}
To train our systems, we synthetically generated data containing eight different
slot types. These are: Time, Date, FoodItem, ListItem, ToDoList, Drink, and Appliance.
See Table~\ref{tab:slots} for examples.
These slot types represent different phrase categories. For example, ListItems are mostly noun phrases and ToDoLists are mostly verb phrases or clauses. In order to generate the data used in our experiments, we use annotated data, which are representative of utterances directed to a commercial SLU system that contains one of the aforementioned slot types and replace them with two or more entities of the same slot type with the coordinating conjunction such as {\it ``and''} between the last entity and the penultimate entity. An utterance example {\it ``set a timer for five minutes''} would become {\it ``set a timer for five and ten minutes.''} 

\begin{table}[t!]
\begin{center}
\begin{tabular}{|l|r|}
\hline \bf Slot type  & \bf Examples \\ \hline
FoodItem & brown sugar, chicken breast  \\
ListItem & shaving cream, bottled water  \\
ToDoList &  call mom, clean windows\\
Drink & coca cola, red bull  \\
Appliance & downstairs light, backdoor camera  \\
\hline\hline
ArtistName &  james brown, kesha \\
EventName & lunch meeting, doctors appointment  \\
MealType &  appetizer, dinner \\
\hline
\end{tabular}
\end{center}
\caption{\label{tab:slots} Examples of various slot types }
\end{table}

From single entity slot annotations, we created slot catalogs and used it to
sample multiple entities of a same slot type. We use BIO tagging scheme to
represent conjunct spans. Table~\ref{data_representation} shows a typical
utterance with BIO tags. We then used annotated data to sample carrier phrases
that contain the relevant slot types and replaced a single-valued slot with
multiple entities sampled from catalog.

\begin{table}[t]
\begin{center}
\scalebox{0.93}{
\begin{tabular}{l|r r r r r r r r}
\hline
tag & O & B-C & I-C & CC & B-C & O & O & O  \\
token  & add & peanut & butter & and & jelly & to & my & list  \\ 
\hline
\end{tabular}}
\end{center}
\caption{\label{data_representation} An utterance with its corresponding labels }
\end{table}

We generate 420k utterances this way. The mean and median utterance length is 10 and 10.5 words respectively. We split our data in 80-10-10 train, validation, and test
splits. 60\% utterances in our data have two conjuncts, 28\% have three
conjuncts, and 12\% have four or more conjuncts. To test whether our model can generalize over unseen slot types, we have created a separate test set that only has coordinated slots that are not present in training data. We chose four slots that were not seen during training: ArtistName, EventName, MealType, and CityName. We use the same procedure to create these four test sets and combine them to form a single test set of unseen slot types. This test set contains 4k utterances in total with 1k utterances for each slot type. We report accuracy and F1 score on this test set with and without adversarial training.

\begin{table*}
  \centering
    \scalebox{0.96}{
    \begin{tabular}{l|r|r|r|r|r|r|r|r|r|r|r|r}
      Exp & Char  & Word  & Loss & \multicolumn{2}{c}{2 entities} &
\multicolumn{2}{c}{3 entities} & \multicolumn{2}{c}{4 or more} & \multicolumn{2}{c}{Latency(ms)}  & Size\\
       & Encoder & Encoder &  & F1 & Acc & F1 & Acc & F1 & Acc & P90 & P99 & mbs\\
    \hline
    1 & None & Random  & Cross Entropy & 0.98 & 0.96 & 0.96 & 0.91 & 0.92 & 0.77 & 2.85 & 3.43  & 64    \\
    2 & None & Random  & +CRF & 0.98 & 0.97 & 0.96 & 0.93 & 0.92  & 0.77 & 3.67 & 4.42  & 64    \\
    3 & None & FastText  & Cross Entropy &0.98 & 0.96 & 0.96  & 0.92 & 0.93 & 0.80 & 2.87 & 3.40   & 64 \\ 
    4 & None & FastText  & +CRF & 0.98 & 0.97 & 0.97 & 0.93  & 0.93 & 0.80 & 3.62 & 4.37  & 64 \\
    \hline
    5 & Random & None & Cross Entropy & 0.97 & 0.95 & 0.94 & 0.88 & 0.89 & 0.72 & 9.03 & 11.33 & 17   \\
    6 & Random & None & CRF & 0.97 & 0.96 & 0.95 & 0.91 & 0.90 & 0.75 & 9.86 & 12.04 & 17   \\
    \hline
    7 & Random & Random  & Cross Entropy &0.98 & 0.97 & 0.96 & 0.92 & 0.92 & 0.78 & 6.38 & 7.89  & 70     \\
    8 & Random & Random  & +CRF & 0.98 & 0.97  & 0.96 & 0.93 & 0.92 & 0.78 &
7.34 & 9.00     & 70   \\
    9 & Random & FastText  & Cross Entropy& 0.98 & 0.97 & 0.96 & 0.92 & 0.93 & 0.80 & 6.47 & 7.73  & 70  \\
    10 & Random & FastText  & +CRF & 0.98   & 0.97 & 0.96 & 0.93  & 0.93 & 0.80 & 7.37 & 9.16 &  70  \\

  \end{tabular}}
  \caption{Various model configurations and results on our in domain test sets. ``None'' means that we did not include that
component in the architecture, Random and FastText refer to random
initialization or FastText initialization of word embeddings.}\label{config}
\end{table*}

\begin{figure}
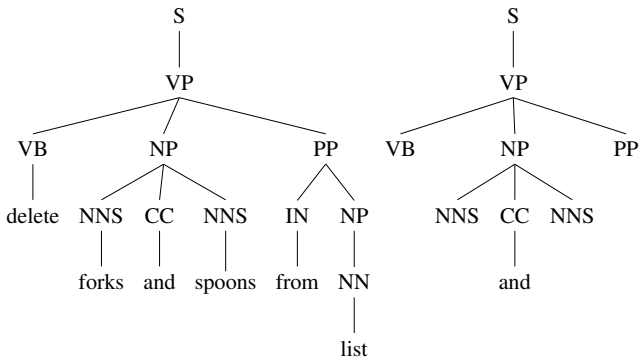

  \begin{center}

\scalebox{0.84}{
    \Tree [.S [.VP [.VB delete ] [.NP [.NNS forks ] [.CC and ] [.NNS spoons ] ]
            [.PP [.IN from ] [.NP [.NN list ] ] ] ] ]
    \Tree [.S [.VP [.VB ] [.NP [.NNS ] [.CC and ] [.NNS ] ]  [.PP ] ] ]
}
    \caption{An example a parse tree of an utterance (left) and tree-pattern
matching the tree (right)}
  \end{center}
\end{figure}
\subsection{Parser}

We also try using a off-the-shelf constituency-based
syntactic parser to parse coordination. Just as our DNN model, we test the parser whether it can reliably find spans of conjuncts. We use popular Stanford parser~\cite{socher2013parsing} to parse our data. Using our training data, we can extract tree patterns that capture coordinating conjunctions and conjuncts. For example, it is obvious that ``CC'' is important category to look for in our task. Our tree patterns are inspired by minimally connected patterns
~\cite{P02-1018}. 
Figure~2 shows a parse of an example sentence and the matching pattern. Any
pattern that occurs less than two times is pruned away. During evaluation, we
use the parser to parse utterances and if a parsed utterance matches the
pattern, we can extract span information by traversing leaves of special nodes
in the pattern. In the example, leaf nodes of the two direct siblings of ``CC''
node, i.e., forks and spoons are the required span conjuncts.

\subsection{Configurations}

We experiment with various choices for the components as shown in Table~\ref{config}. We use bi-directional LSTMs for both character and word-level encoders and a fully connected layer as our decoder. To test models for sizes and latency, we experiment with removing word and character-level encoders as well as various losses for training.

For the best performing models, we also experiment with adversarial loss to test
whether it helps with generalization capability of our models. For this
experiment we take our configurations 4 and 10, i.e., FastText word embeddings
$+$ CRF and character embeddings $+$ FastText word embeddings $+$ CRF and add a
gradient reversal layer~\cite{ganin2016domain} to make the model more slot-type
agnostic. In both configurations we add a task-specific LSTM layer, keeping the base LSTM layer as such. We expect base LSTM layer to learn features that are common to both sequence tagging and slot classification task. For a sentence $S = {x_{1,2...T}}$, where $x_i$ is the input representation of a word, we get $h_1, h_2...h_t = BiLSTM_{base}(x_{1,2...T})$. We pass this to another bi-directional LSTM layer, $g_1, g_2...g_t = BiLSTM_{seq}(h_1, h_2...h_t)$, the $g_i$ is used as the output for the sequence tagging task. In the adversarial setting, we also use last output of the $BiLSTM_{base}$ to predict slots after passing it through a gradient reversal layer (GR) that is $d = GR(h_t)$ which is passed to an affine layer and softmax to predict slots.

We tune our training-related hyper parameters on a simpler model and keep them
constant for all configurations. We use a dropout~\cite{srivastava2014dropout}
of 0.5 on the embedding layer and a variational
dropout~\cite{kingma2015variational} of 0.01 between time-steps for LSTM. We use
ADAM~\cite{kingma2014adam} for our optimization method with learning rate of
$0.002$\@. Our character and word embedding sizes are 100 and 300 respectively.
For hidden dimension sizes, we choose 100 for character bi-directional LSTM and
200 for word bi-directional LSTM. We test convergence on our development set for
early stopping. We observe the validation accuracy and terminate the training when the accuracy does not improve for a fixed number of steps, which is set to 10 for our experiments. The latency results are for inference time on CPU.

We had to tune our $\gamma$ carefully as model performance depended heavily on
the value of $\gamma$ when adversarial training was used. $\gamma$ was chosen to
be $2$ after experiments on the development set.

\begin{table}
  \centering
    \scalebox{0.9}{
    \begin{tabular}{l|r|r}
    Experiment    & F1 & Acc    \\
    \hline
    Base  & 0.60 & 0.38    \\
    $+$Adversarial  & 0.69 ($+$15\% rel) & 0.48($+$26\% rel)   \\
    \hline \hline
    In-domain Test Set (Base) & 0.98 & 0.97 \\
    In-domain Test Set ($+$Adv) & 0.98 & 0.97 \\ 
  \end{tabular}}
  \caption{Generalization experiment results on a different test distribution
(tests sets with slot types that are not present in our training data). We used
our most accurate model (character embeddings $+$ FastText word initialization $+$ CRF
loss) along with adversarial training for these experiments. The first two rows
show that adversarial training helps our model to perform better on the
out-of-domain test set. The third and fourth rows show that adversarial training
does not make our model degrade on in-domain test set.}\label{unseen}
\end{table}

\section{Results}\label{sec:res}

For results, we report accuracy and F1 score. For accuracy, we only consider the
exact matches of all conjuncts to be correct in an utterance. Utterances with
any mislabeled tokens are given a score of 0\@. Note that for F1 score, we
compute precision and recall per conjunct rather than per token. Thus, an entire
span of conjunct need to be labeled correctly to count.

Our off-the-shelf parser baseline did not perform well. At conjunct-level,
precision and recall was only 0.31 and 0.26 respectively for noun phrases and
nouns. For verb phrases corresponding numbers were slightly better. It was 0.49
for precision and recall was 0.64\@. We attributed one of the reason for
parser's poor performance to the absence of punctuation in spoken language. We
verified this hypothesis by manually adding punctuation to a small subset. On
this subset, the parser had much better accuracy. Our test set contains coordination of more than two conjuncts without any punctuation. The parser especially struggled with such utterances.

We show our results on 2, 3, and 4 or more coordinated entities (conjuncts) on the test set in Table~\ref{config}.

In order to compare latency, we report p90 (90\% of utterances can be processed under this time) and p99 (99\% of utterances can be processed under this time) times to on our development set at inference time on CPU. We also measure the size of our models.  To test the generalization abilities of the system, we create a test
set on a new slot type which the model has not seen while training (See Section~\ref{sec:data} for details). We present results on the unseen test set for our best models in Table~\ref{unseen}.

\begin{table*}[th]
\begin{center}
\scalebox{0.93}{
\begin{tabular}{l|r r r r r r r r r}
\hline
Tokens  & what's & the & forecast & for & santa & clara & and & boston & \\
Ground Truth & O & O & O & O & B-C & I-C & CC & B-C & \\
\hline
No adv loss & O & O & B-C & I-C & I-C & I-C & CC & B-C &  \\
Adv loss & O & O & O & O & B-C & I-C & CC & B-C & \\
\hline\hline
Tokens  & add & pokemon & go & and & bed & to & the & calender & today  \\
Ground Truth & O & B-C & I-C & CC & B-C & O & O & O & O  \\
\hline
No adv loss  & O & B-C & B-C & CC & B-C & O & O & O & O   \\
Adv loss &  O & B-C & I-C & CC & B-C & O & O & O & O   \\

\end{tabular}}
\end{center}
\caption{\label{example} Examples utterances where adversarial training helps}
\end{table*}

We note a few general trends from Table~\ref{config}.
First, irrespective of the network configuration, performance degrades as we
increase the number of entities. This is well expected since increasing the
number of conjuncts exponentially increases the number of possible segmentation
points making it harder for the model to learn all the associations with limited data.
Next, we observe that adding a CRF layer gives us improvements in accuracy across the board but it comes with a latency penalty as well, which is expected due to additional computation at the CRF layer. Compared to our fastest models, we see an increase in latency of almost 30\% when CRF layer is added. Next, we observe that adding FastText initialization for word embeddings helps when we only have word embeddings but when we add in the character embeddings the effect vanishes.

We suspect this is because FastText includes character information and probably
helps unseen and under-represented words in the training data but having
character encoder had similar impact. Our smallest model is almost 25\% of our
largest models and our fastest models being almost four times faster than our
slowest models. We also observe that there is a trade-off of memory versus
latency when only character embeddings are employed. There is only a small accuracy hit if only character embeddings are employed.

Our most accurate model on in-domain test set is the one with character-level
encoder, word-level encoder as well as the CRF layer. But if we take both accuracy and latency into account, we observe that by just using word embeddings with FastText initialization, we get very similar performance while being much faster.

As seen in Table~\ref{unseen}, by adding adversarial loss to our model training, we observe a gain on our out-of-domain test set while the performance on the in domain test set remains the same. This shows that adversarial training increases generalization capability of our models. We observe a 15\% relative gain over baseline on F1 score and 26\% relative gain on accuracy. We provide some examples of sentences where adversarial training improves accuracy in
Table~\ref{example}.

\section{Related Work}\label{sec:related}

A way to add structure to language analyses without relying on full
tree-producing parsers is to use shallow parsing. Shallow parsing problem is
well-established subfield of NLP. There have been many shared tasks on this subject such as CoNLL-2000 shared task on chunking~\cite{tjong2000introduction}. Solving shallow parsing as sequence tagging is also well-established. Our task is specific type of shallow parsing that finds coordinating structures. There are several earlier works that have tried to do the same with and without using machine learning. \cite{agarwal1992simple} named the task conjunct identification. \cite{voutilainen1995nptool} tried to solve this task by writing manual grammars. Statistical approaches used in the past include transformation-based learning~\cite{ramshaw1999text}, which is the same technique used by Brill Tagger~\cite{brill1992simple}, memory-based learning~\cite{daelemans1999memory, argamon1998memory}, and support vector machines~\cite{pradhan2004shallow}. Use of sequence tagging techniques such as HMMs~\cite{molina2002shallow} and CRFs~\cite{sha2003shallow} has also been popular for similar tasks. 

Since popularization of DNN usage in NLP, RNN-based models have replaced CRF as the de-facto standard for sequence tagging models~\cite{ling2015not, ma_hovy_2016,lample2016neural}. Many models employ RNN with CRF loss~\cite{ma_hovy_2016, lample2016neural}. More recently, non-recurrent models such as CNN~\cite{xu2013convolutional,
xu2018double,cheng2017cnn} and transformer network~\cite{vaswani2017attention} have become popular for sequence labeling tasks due to their efficiency.

Character-based embeddings have been found effective for wide-range of NLP tasks including parsing \cite{kitaev2018constituency}. These embeddings are good at learning morphology of words, replacing POS-based features in many tasks. They also makes the models more robust to unknown words. \cite{ling2015not} introduced character-based embeddings that is used in our setup and \cite{lample2016neural} is perhaps the most similar sequence labeling architecture to ones used for this work. More recent breakthroughs in dynamic word representation also includes character models of words~\cite{peters2018deep}. 

Adversarial training has gained a lot of popularity in recent times as evidenced by ever-increasing papers on generative adversarial networks. Most of the early work on adversarial networks and adversarial training has been in the vision community. However, more recently there are examples of adversarial training applied to NLP. \cite{wu2017adversarial,he2016dual} use adversarial training to train machine translation models.\cite{li2017adversarial} uses adversarial training for dialog generation.

\section{Conclusion}\label{sec:conclusion}

Parsing coordination structure may seem like a solved problem using off-the-shelf syntactic parsers. However, spoken language is less structured than written texts and ASR outputs lack important parsing cues such as punctuations. Annotated real-world spoken data to train coordination parsing models is also hard to come by. In this work, we show that we can train a relatively simple neural chunking model that achieves good performance on this task using only synthetic data. We experiment with different network configurations and show that adding a CRF layer consistently provides higher accuracy in such tasks. By analyzing model size and inference latency under different network configurations, we find that word embedding based models are useful in latency constrained environments while character embedding based models are useful in memory constrained ones. Furthermore, we show that these models achieve better generalization with addition of adversarial loss during training.

In real-world scenarios, the parser we presented can be integrated to a
slot-type detection module to build a two-tiered SLU system. The parser and the
slot-type detection module can run in parallel and the results can be merged
through simple rules. To further improve run-time efficiency of such system, a
classifier can be added that can detect coordinated intents or entities and only
the ones with coordination are run through the parser.  

We hope this work will provide meaningful insights to practitioners working with
SLU systems on how they can parse coordination in spoken language.

\bibliographystyle{IEEEbib}
\bibliography{refs}

\end{document}